\documentclass[conference]{IEEEtran}
\IEEEoverridecommandlockouts
\usepackage{graphicx}
\usepackage{amsmath,amsthm,amsfonts,amssymb,xfrac}
\usepackage{mathtools}
\usepackage{booktabs}
\usepackage{siunitx}
\usepackage{multirow}
\usepackage{hyperref}
\usepackage{cleveref}
\usepackage{cite,soul,bm}
\usepackage{xcolor}
\usepackage[table,xcdraw]{xcolor}

\newtheorem{remark}{Remark}

\newcommand{\T}{\top}

\newif\ifshowcomments
        \showcommentstrue  
    \ifshowcomments 
        \newcommand{\bae}[1]{\hl{[SB: #1]}\protect\color{black}} 
        \newcommand{\di}[1]{\hl{[DI: #1]}\protect\color{black}}
        \newcommand{\ft}[1]{\hl{[FT: #1]}\protect\color{black}} 
        \newcommand{\jd}[1]{\hl{[JD: #1]}\protect\color{black}}
        \newcommand{\avi}[1]{\hl{[AS: #1]}\color{black}}
    \else
        \newcommand{\bae}[1]{}
        \newcommand{\di}[1]{}
        \newcommand{\ft}[1]{}
        \newcommand{\jd}[1]{}
        \newcommand{\avi}[1]{}
    \fi

\title{ONRAP: Occupancy-driven Noise-Resilient Autonomous Path Planning}
\author{Faizan M. Tariq$^{1}$, Avinash Singh$^{1}$, Vipul Ramtekkar$^{2}$, Jovin D'sa$^{1}$, \\David Isele$^{1}$, Yosuke Sakamoto$^{1}$, Sangjae Bae$^{1}$
\thanks{$^{1}$ Honda Research Institute, CA, USA.
Email: \texttt{\{faizan\_tariq, avinash\_singh, jovin\_dsa, disele, yosuke\_sakamoto, sbae\}@honda-ri.com}}
\thanks{$^{2}$ Honda Research and Development, Japan. Email: \texttt{ramtekkar\_vipul@ jp.honda}}
}

\begin{document}
\maketitle

\begin{abstract}
Dynamic path planning must remain reliable in the presence of sensing noise, uncertain localization, and incomplete semantic perception. We propose a practical, implementation-friendly planner that operates on occupancy grids and optionally incorporates occupancy-flow predictions to generate ego-centric, kinematically feasible paths that safely navigate through static and dynamic obstacles. The core is a nonlinear program in the spatial domain built on a modified bicycle model with explicit feasibility and collision-avoidance penalties. The formulation naturally handles unknown obstacle classes and heterogeneous agent motion by operating purely in occupancy space. The pipeline runs in real-time (faster than 10 Hz on average), requires minimal tuning, and interfaces cleanly with standard control stacks. We validate our approach in simulation with severe localization and perception noises, and on an F1TENTH platform, demonstrating smooth and safe maneuvering through narrow passages and rough routes. The approach provides a robust foundation for noise-resilient, prediction-aware planning, eliminating the need for handcrafted heuristics. The project website can be accessed at https://honda-research-institute.github.io/onrap/
\end{abstract}

\begin{IEEEkeywords}
Autonomous Driving, Model Predictive Control, Collision Avoidance, Occupancy Flow, Path Planning
\end{IEEEkeywords}

\section{Introduction}
Autonomous vehicles (AVs) and mobile robots must plan paths in dynamic and uncertain environments, with strict computational constraints. In practical deployments, localization drift, sensor dropouts, inconsistent detections, and incomplete semantics are the norm rather than the exception. Under these conditions, a path planner must generate kinematically feasible, smooth, and safe trajectories while reacting promptly to evolving obstacles and traffic flow. Failures in this layer cause unnecessary braking, erratic lateral behavior, or, in the worst case, collisions and deadlocks. This paper addresses the problem of \emph{dynamic, noise-resilient local path planning} with minimal assumptions about object identities or intent, relying instead on occupancy representations \cite{fleuret2008multicamera} that inherently tolerate ambiguity and partial observability.

\begin{figure}
    \centering
    \includegraphics[width=0.8\columnwidth]{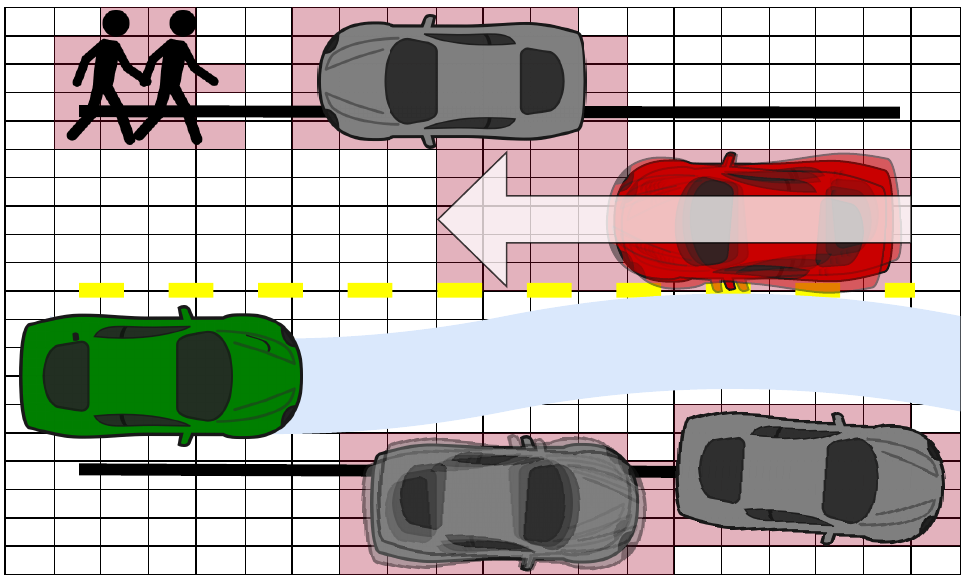}
    \caption{\textbf{Motivation.} The ego vehicle (green) must deviate from the lane center to avoid parked vehicles (gray) along the roadside while remaining aware of oncoming traffic (red). Blurred vehicles indicate perception uncertainty. The planner processes the occupancy grid, with occupied cells highlighted in red, and generates the blue trajectory, enabling safe navigation around obstacles and dynamic traffic, in the presence of real-world uncertainties.}
    \label{fig:motivation}
\end{figure}

Object-centric planning pipelines \cite{sadigh2018planning,tariq2022slas} often depend on reliable object classification and tracking to reason about future motion. While effective in curated settings, they degrade when upstream modules mislabel agents or intermittently miss detections. 
Heuristic safety margins can only partially mitigate this brittleness; excessively large margins reduce efficiency, yet still fail in dense or occluded scenes.
On the other hand, raw-grid methods, which plan directly on rasterized occupancy maps \cite{stentz1995focussed,koenig2002dlite,dolgov2008hybrida}
are robust to category errors, but are sometimes criticized for losing kinematic structure or for yielding overly conservative paths when predictions are absent or uncalibrated. Bridging these extremes requires a formulation that natively consumes occupancy information directly \emph{and} preserves vehicle kinematic feasibility, while remaining sufficiently light-weight for real-time deployment. 

Consider the scenario depicted in \autoref{fig:motivation}. The ego vehicle must smoothly maneuver around both static and dynamic obstacles under perception uncertainty, while ensuring that the resulting path remains kinematically feasible. An occupancy grid-based planner, where each cell’s occupancy can be individually specified, enables straightforward handling of real-world uncertainties by marking affected cells accordingly. This design ensures that the planner’s computational complexity is independent of the number or shapes of obstacles and depends solely on the dimensions of the input grid.

We introduce an occupancy-driven noise-resilient autonomous path planner (ONRAP) that operates in the spatial domain. ONRAP builds on a modified bicycle model \cite{kong2015kinematic} and encodes feasibility and collision avoidance directly through occupancy sets. A key design choice is to represent both static and dynamic agents through occupancy grids, optionally augmented with occupancy-flow predictions. By eliminating reliance on object identity or intent labels, ONRAP remains robust to perception noise and semantic uncertainty. At the same time, its spatial-domain nonlinear program (NLP) respects curvature and control limits, producing smooth, trackable paths for the downstream modules. In essence, ONRAP combines the \emph{class-agnostic robustness} of occupancy-based methods with the \emph{structure and performance} of kinematic optimization.

We evaluate the performance of ONRAP in two deployment contexts: (i) simulation with severe noise injections and challenging map geometries (high curvature, narrow passages, and occlusions), and (ii) a real-world F1TENTH testing platform \cite{o2020f1tenth} for empirical assessment of timing and robustness. In simulation, we stress-test the planner by perturbing localization and perception data with stochastic outliers, and quantifying collision rate, minimum obstacle clearance, path smoothness, and solve-time distributions. On F1TENTH, we evaluate responsiveness and reliability under limited compute, while examining the effects of occupancy-flow guidance on safety and replanning stability in dynamic traffic scenes.

\subsubsection*{Contributions} 
This paper makes two primary contributions:
\begin{enumerate}
    \item We present an occupancy-driven noise-robust planning framework that formulates path generation as a spatial domain NLP constrained directly by occupancy and optional occupancy-flow fields. This formulation yields kinematically feasible, smooth, and class-agnostic trajectories in real-time.
    \item We conduct comprehensive evaluations in a noisy simulation environment and on F1TENTH hardware platform, demonstrating reduced collisions and a smooth trajectory; occupancy-flow guidance improves safety margins and temporal stability in a dynamic environment. 
\end{enumerate}

\textbf{Why this matters:} Real-world AV and mobile robot deployment requires planners that degrade gracefully when the upstream modules are compromised. By utilizing occupancy as the primary safety signal and incorporating flow-based guidance, our method sidesteps brittle dependencies on semantic correctness while retaining the benefits of structured, kinematics-aware optimization. The resulting planner is simple to implement, computationally efficient, and adaptable to a variety of perception stacks, making it a strong candidate for both research baselines and practical integration.


\subsection*{Related Work}\label{sec:related}
\textbf{Occupancy grids and occupancy \emph{flow}.} 
Classical grid-based planners such as A$^\star$, D$^\star$, and Hybrid A$^\star$ reason over rasterized free space and often apply post hoc smoothing to restore kinematic feasibility; they work well with static or slowly varying maps but struggle when free space evolves quickly due to moving agents \cite{koenig2002dlite,dolgov2008hybrida}. To better capture dynamics, recent prediction works model space as time-varying occupancy with per-cell motion, e.g., \emph{Occupancy Flow Fields} that jointly predict occupancy and 2D flow vectors \cite{mahjourian2022occupancy}, and unified perception–prediction models that implicitly represent occupancy/flow over time \cite{agro2023implicit}. Instance-centric BEV predictors like FIERY target future instance segmentation and motion fields from cameras \cite{hu2021fiery}. Most planners that consume these predictions, however, use them as \emph{soft costs} or priors rather than feasibility constraints, which can yield conservative or inconsistent behavior under severe noise or occlusion. \textbf{ONRAP differs} by taking in class-agnostic occupancy (and optional occupancy-flow) and \emph{embedding them directly as constraints} in a spatial-domain trajectory optimization, yielding kinematically feasible paths that maintain progress even when semantics are incomplete or unstable.

\textbf{Learning-based planners and generalization limits.} Imitation- and learning-based planners that rasterize scenes (lanes/agents) have shown compelling closed-loop results (e.g., ChauffeurNet) but typically rely on stable \emph{semantic} inputs (lane graphs, instance tracks, intent cues, etc.) at train and test time \cite{bansal2019chauffeurnet}. In practice, these semantics can be brittle across different operation domains and distribution shifts; scaling requires heavy labeling, curated simulations, and careful dataset generation. Even occupancy/flow predictors used as learned priors often assume consistent semantics for training targets (e.g., agent masks, lane layers) and can degrade when those signals are partial or noisy \cite{hu2021fiery,mahjourian2022occupancy}.

\emph{ONRAP} adopts a semantics-agnostic design, grounding collision avoidance directly in occupancy and flow representations that modern perception modules can generate even when instance tracking or classification is unreliable, while maintaining a lightweight and interpretable planning core.
Relative to grid search baselines \cite{koenig2002dlite, dolgov2008hybrida}, ONRAP reasons over evolving free space without hand-tuned heuristics, and compared to learning-only stacks \cite{bansal2019chauffeurnet}, it provides explicit feasibility handling and clear safety margins derived from the occupancy constraints themselves.

This paper takes inspiration from~\cite{tariq2025frenet}, which formulated spatial-domain bicycle kinematics to solve a nonlinear program (NLP) for optimal path planning along a prescribed reference. While effective in structured settings,~\cite{tariq2025frenet} had four key limitations: (i) reliance on a smooth, consistent reference path; (ii) curvature-coupled action bounds that could render feasible motions infeasible; (iii) repeated Cartesian$\leftrightarrow$Frenet coordinate transformations within the optimization loop; and, (iv) dependence on semantic information (agent classes/boxes or risk maps) to encode safety. We remove these bottlenecks by: (i) formulating the NLP in an ego-anchored spatial domain without requiring a smooth reference, making it robust to unmapped areas, lane shifts, and construction scenes; (ii) making action bounds reference-invariant by defining curvature/steering in the ego frame; (iii) operating natively in an ego-centric BEV/occupancy frame, avoiding global$\leftrightarrow$ego conversions; and, (iv) enforcing semantics-free safety via occupancy and optional occupancy-flow risks.

\begin{figure*}
    \centering
    \includegraphics[width=0.95\linewidth]{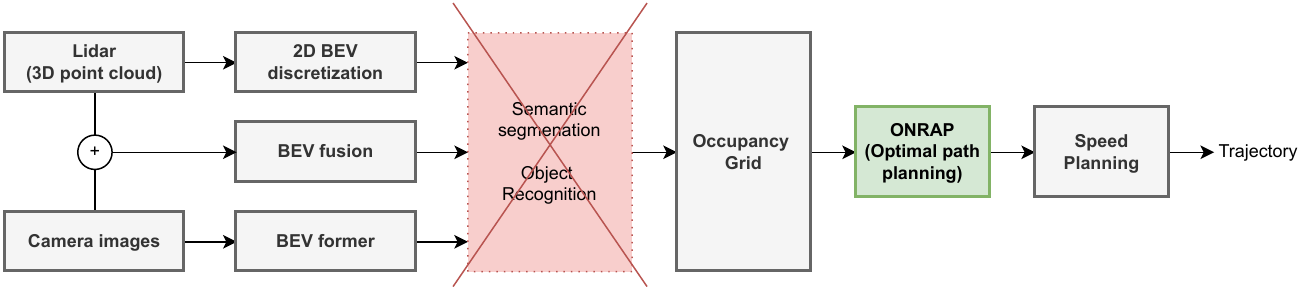}
    \caption{\textbf{Pipeline.} Raw sensory input data is processed by the BEV generation modules to produce BEV data for the occupancy grid generator, eliminating the need for intermediary semantic segmentation and object recognition modules. The resulting occupancy grid is then fed into our ONRAP module, which computes a kinematically feasible path for the downstream speed planning module.}
    \label{fig:pipeline}
\end{figure*}

\section{Method}\label{sec:method}
\autoref{fig:pipeline} provides an overview of the mobility pipeline, which receives raw sensory input and produces a trajectory for the downstream controller to follow. In this work, we omit semantic segmentation and object recognition from the pipeline and propose a scheme for optimal path planning using arbitrary noisy occupancy grids. This section details the key components of our approach, including the underlying kinematic model, occupancy grid processing, reference path generation, constraint formulation, and the resulting nonlinear programming (NLP) problem.

\subsection{Spatial domain bicycle kinematics} In our previous work, we transformed the bicycle kinematics model from the space-time domain to the space-only domain, using a fixed longitudinal step, which was the key to simplifying the collision avoidance set. To recap, for each longitudinal step $k$, the kinematics model reads:
\begin{align}
    x_{k+1} &= x_k + \Delta s,\label{eq:dyn_x}\\ 
    y_{k+1} &= y_k + \frac{\sin(\psi_k+\beta_k)}{\cos(\psi_k+\beta_k)}\Delta s\nonumber\\&=y_k + \tan(\psi_k+\beta_k)\Delta s,\label{eq:dyn_y}\\
    \psi_{k+1} &= \psi_k + \frac{\Delta s}{\ell_r}\frac{\sin\beta_k}{\cos(\psi_k+\beta_k)},\label{eq:dyn_heading}
\end{align}
where $\{x,y,\psi\}$ denotes $\{$x-coordinate, y-coordinate, heading$\}$, $\beta_k \approx \frac{l_r}{l_f+l_r}\delta_k$, and $\Delta s$ is a fixed longitudinal step size. It is important to note that this kinematics model is well defined within the range $\psi_k+ \beta_k \in (-\pi/2,\pi/2)$ and the approximation error deteriorates the kinematic feasibility near the bounds (i.e. $\pm\pi/2$).

\begin{remark} While this kinematic model was originally derived for the Frenet frame, it is directly applicable to any subset of the Cartesian frame where $\psi_k + \beta_k \in (-\pi/2, \pi/2)$ holds. In practice, this includes ego-centric planning formulations that reinitialize the local frame at each solution instance with zero heading (i.e., $\psi_0 = 0$), ensuring that the mapping remains well-conditioned over the planning horizon.
\end{remark}

\subsection{Occupancy grid handling}
\begin{figure}
    \centering
    \includegraphics[width=1\columnwidth]{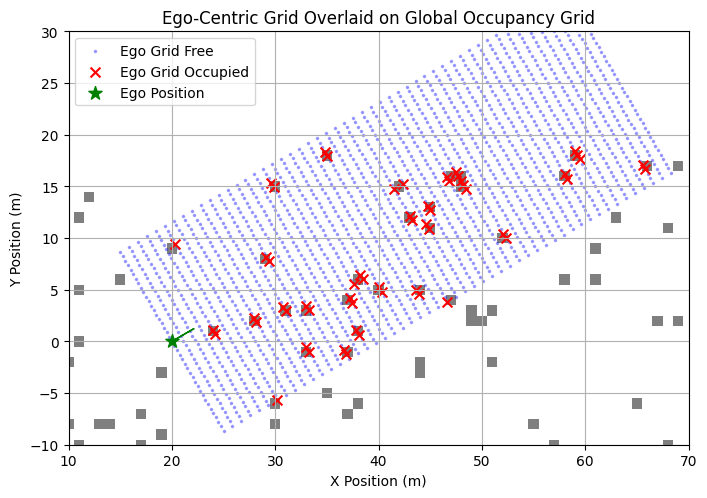}
    \caption{\textbf{Ego-centric occupancy grid.} The ego-centric occupancy grid is represented as a fixed-size matrix centered on the ego vehicle. Arbitrary occupancies from the global map are projected into this local frame, preserving spatial relationships while ensuring constant computational complexity regardless of obstacle count or shape.}
    \label{fig:ego-centric-occupancy-example}
    \vspace{-10pt}
\end{figure}

\textbf{Ego-centered occupancy grid:} Let the global occupancy grid provided by the preceding Perception module at time $t$ (see \autoref{fig:pipeline}) be denoted as $\mathcal{G}_t^p$. We define a fixed-size, parametric ego-centric occupancy grid at time $t$ (omitted for simplicity) as $\mathcal{G} \in \mathbb{R}^{N_w \times N_l}$, where $N_w$ and $N_l$ represent the grid width and length, respectively. The global grid $\mathcal{G}_t^p$ is projected onto the ego-centric grid $\mathcal{G}$, as illustrated in \autoref{fig:ego-centric-occupancy-example}. Notably, under this transformation, the computational complexity of the ego-centric representation depends only on the grid dimensions rather than the number or shape of obstacles, ensuring consistent computational complexity for the nonlinear program in \autoref{sec:nlp}. This highlights the advantage of an image-based representation for planning.

\textbf{Occupancy flow prediction:} Any occupancy in the ego-centric grid is treated uniformly, with no distinction between static and dynamic obstacles. The temporal evolution of occupancies is addressed by predicting occupancy flow \cite{niemeyer2019occupancy} based on observed grids. To keep the model lightweight and real-time implementable, we employ a simple Kalman filter to estimate per-cell flow $(v^x, v^y)$ in \emph{cells/frame}, given two consecutive (arbitrary) occupancy grids $O_{t-1}, O_t \in [0,1]^{H \times W}$. Grid coordinates follow the image convention: row $i$ (down) and column $j$ (right). Consequently, $v^y$ affects rows, while $v^x$ affects columns.
For each cell $(i,j)$ and each axis $\alpha \in \{x,y\}$, we maintain a velocity estimate $v^\alpha$ and its associated variance $p^\alpha$ as follows.
\begin{align}
v_t^\alpha &= v_{t-1}^\alpha + w^\alpha, \quad w^\alpha \sim \mathcal{N}(0,q), \\
z_t^\alpha &= v_t^\alpha + n^\alpha, \quad n^\alpha \sim \mathcal{N}(0,r),
\end{align}
where $q$ is the process noise variance and $r$ is the measurement noise variance. 
The measurement $z_t^\alpha$  is obtained through a $3{\times}3$ local search following these steps: (i) Predict the integer landing index
$(\hat{i},\hat{j})=\Big(\mathrm{round}(i+v^y\Delta t),\,\mathrm{round}(j+v^x\Delta t)\Big)$, clamp it to the grid, and search the neighborhood $\{(\hat{i}{+}\delta_i,\hat{j}{+}\delta_j)\mid \delta_i,\delta_j\in\{-1,0,1\}\}$ in $O_t$; (ii) Choose the  cell with the highest occupancy and let $(\Delta i^*,\Delta j^*)$ denote its displacement from $(i,j)$; and (iii) Set $z_t^y=\Delta i^*/\Delta t, z_t^x=\Delta j^*/\Delta t$.
For active cells $\mathcal{A}=\{(i,j):O_{t-1}(i,j)\ge\beta\}$, we perform the Kalman updates as follows:
\begin{align}
p^\alpha &\leftarrow p^\alpha + q, \quad
K^\alpha \leftarrow \frac{p^\alpha}{p^\alpha + r}, \\
v^\alpha &\leftarrow v^\alpha + K^\alpha (z^\alpha - v^\alpha), \quad
p^\alpha \leftarrow (1{-}K^\alpha)p^\alpha,
\end{align}
then clip $v^\alpha \in [-v_{\max}, v_{\max}]$ and floor $p^\alpha \ge \epsilon$. $\beta$ is the occupancy threshold, and we set it to $0.5$. The estimated per-cell flow $(v^x,v^y)$ is then used to linearly predict occupancy grids in any time (i.e., constant velocity prediction). 

For the F1TENTH experiments in \autoref{sec:f1tenth}, the per-cell flow $(v^x,v^y)$ at each frame $t$ is smoothed out using a moving average filter, i.e., $\bar{v}^x_t = \text{mean}(v^x_t,v^x_{t-1},\ldots)$, which improves the consistency of occupancy-flow predictions, particularly under high localization and perception noise. Note that ONRAP is agnostic to the choice of flow prediction models; therefore, more sophisticated approaches such as those in \cite{mahjourian2022occupancy, agro2023implicit} can be readily integrated.

\textbf{Occupancy risk} We directly incorporate the occupancy risk into the objective function as a soft constraint\footnote{While the soft constraint does not guarantee collision avoidance, it preserves recursive feasibility across different solution instances. Moreover, under the path–speed decomposition scheme, the downstream speed planner is responsible for maintaining a safe longitudinal distance.}. Formally, the occupancy risk is defined as:
\begin{equation}
    r_\mathcal{G} = \sum_{k}\;\mathcal{G}_k^\T f_\text{grid}(y_k),\label{eq:cost_grid}
\end{equation}
where $f_\text{grid}(\cdot)$ is a vectorized risk function that maps the decision variable $y_k$ to the cost at each element in $\mathcal{G}_k$. Each $k$-th column, i.e., $\mathcal{G}_k$, is a vector of indicators, i.e.,
\begin{equation}
    \mathcal{G}_{k\mid i} =
    \begin{cases}
      1, & \text{if occupied at row } i, \\
      0, & \text{otherwise.}
    \end{cases}
  \end{equation}

\begin{figure}
    \centering
    \includegraphics[width=1\linewidth]{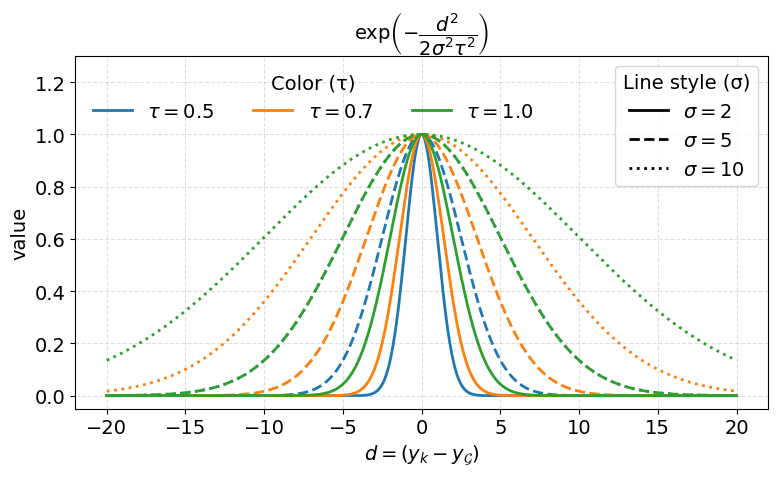}
    \caption{\textbf{Pareto analysis} of the risk function $\exp{\left(-\frac{(y_k-y_{\mathcal{G}})^2}{2(\sigma\cdot\tau)^2}\right)}$ for different $(\sigma,\tau)$ pairs. Here, $\sigma$ represents the desired safety distance, while $\tau$ controls the sharpness of the risk profile (smaller $\tau$ yields a steeper decay).}
    \label{fig:risk_pareto}
\end{figure}

The risk function $f_\text{grid}(\cdot)$ is designed to impose a higher penalty on obstacles that are closer to the ego vehicle. To achieve that, several formulations are possible, including linear, quadratic, or exponential functions. In this work, however, we adopt a modified Gaussian function parameterized by a scaling factor $\tau$ (applied to $\sigma$). Formally, it is defined as:
\begin{equation}
    f_\text{grid}(y_k) = \exp{\left(-\frac{(y_k-y_{\mathcal{G}})^2}{2(\sigma\cdot\tau)^2}\right)},
\end{equation}
where $y_{\mathcal{G}}$ denotes the vector of $\{$y-coordinates$\}$ in the ego-centric grid $\mathcal{G}$. This function facilitates the incorporation of a desired safety distance $\sigma$ to maintain clearance from occupied cells, while smoothly attenuating the risk beyond that distance. Its maximum value is bounded by 1, which simplifies the regularization against other cost terms. The scalar $\tau$ controls the sharpness of the curve, and a sensitivity analysis for different $\tau$ values is shown in \autoref{fig:risk_pareto}.

\begin{remark} Mathematically, $\tau$ and $\sigma$ could be combined into a single term in the risk function without altering its overall behavior. However, defining them separately greatly facilitates practical fine-tuning. For a given safety distance $\sigma$, setting $\tau = 0.7$ yields a risk value of approximately $0.35$ at $d = \sigma$, with negligible risk beyond $2\sigma$; this configuration is leveraged in our validation studies in \autoref{sec:validation}.
\end{remark}

\subsection{Reference path generation given a single point}\label{sec:ref-generation}
Given a deterministic local goal point at each solution instance, intermediary reference waypoints need to be generated for the planner. Unlike existing work \cite{tariq2025frenet}, we allow the reference path to be noisy and inconsistent between solution instances, while allowing high curvatures that may not be kinematically feasible. Several options exist for path generation, e.g., cubic splines \cite{ryu2025iann} and clothoids \cite{ionut2022spherical}, but we opt for the Quintic Hermite spline approach \cite{dougherty1989nonnegativity} due to its precise boundary control, allowing us to enforce start and end conditions on $\{x, y, \theta\}$, and its inherent smoothness.

\begin{remark}
Relaxing the requirement for a smooth and consistent reference path improves robustness to noise, localization drift, and map imperfections. This flexibility allows the planner to re-optimize for feasibility and safety whenever the reference path jitters or demands infeasible curvatures, thereby enhancing adaptation to real-world conditions.
\end{remark}

\subsection{Constraints and bounds}
Existing work \cite{tariq2025frenet} exploited semantic information for static obstacles and road structure to derive explicit collision-avoidance constraints. In this paper, we remove that prerequisite: all obstacles are represented in a single occupancy grid, without object classification or recognition. This simplification reduces constraint complexity but, in exchange, weakens formal collision-avoidance guarantees by relying instead on occupancy-based safety.

We introduce a simple yet important modification to the bounds by adding a heading angle constraint at each planning step, i.e. $\psi_k \in  [\psi_\text{min}, \psi_\text{max}]$, where
\begin{align}
    \psi_\text{min} &> -\pi/2-\beta_\text{min}+\epsilon,\\
    \psi_\text{max} &< \pi/2-\beta_\text{max}-\epsilon,
\end{align}
with a buffer parameter $\epsilon$. This is to ensure that the spatial domain bicycle kinematics \cref{eq:dyn_x}-\eqref{eq:dyn_heading} remain well-defined. Alternatively, inequality constraints can be designed as:
\begin{align}
    \psi_k + \beta_k &> -\pi/2 +\epsilon,\\
    \psi_k + \beta_k &< \pi/2 +\epsilon,
\end{align}
which allows for more precise exploration in the feasible set, at the cost of added complexity. Note that this constraint must not include any slack variable that permits violations; otherwise, kinematic feasibility may not be maintained. 

\subsection{Complete NLP Formulation}   \label{sec:nlp}
The complete nonlinear programming formulation reads:
\begin{align}
    \min_{u}\;\; &(y-y_\text{ref})^\top Q_d (y-y_\text{ref}) + \lambda_\text{grid} r_\mathcal{G} \nonumber\\&+ u^\top Q_u u + \lambda_\text{curve}\sum_k\tan^2 u_k\label{eq:obj} \\
    \text{subject to:}\nonumber\\
    x_{k+1} &= x_k + \Delta s,\\
    y_{k+1} &= y_k + \tan(\psi_k+u_k)\Delta s,\\
    \psi_{k+1} &= \psi_k + \frac{\Delta s}{\ell_r}\frac{\sin u_k}{\cos(\psi_k+u_k)}\\
    y^\text{lb}_{t_k} &\leq y_k \leq y^\text{ub}_{t_k} \label{eq:bounds}\\
    x_0&=\hat{x}_t,~y_0=\hat{y}_t,~\psi_0=\hat{\psi}_t\\
    u_k^\text{min} &\leq u_k \leq u_k^\text{max} 
    \\
    \psi_\text{min} &\leq \psi_k \leq \psi_\text{max},\label{eq:control_bounds}
\end{align}
where operator $\hat{\cdot}_t$ denotes the measured state at time $t$, and the control input is defined as $u = [u_0, \ldots, u_{N-1}]$ over a planning horizon of length $N$. The objective function in \cref{eq:obj} penalizes the following, in order: (i) deviations from the reference path $y_\text{ref}$, (ii) accumulated risk over the occupancy grid $\mathcal{G}$, (iii) control effort, and (iv) accumulated curvature. The constraint in \eqref{eq:bounds} limits lateral exploration, which can be configured to reflect predefined road boundaries (if available), or set to arbitrary bounds as needed.

\begin{remark}
The objective function in \cref{eq:obj} is \textbf{non-convex}, due to the sum of Gaussian terms in \cref{eq:cost_grid}. Empirically, the nonlinear program yields consistent, high-quality solutions when the lateral search range is bounded within $5\sigma$, i.e., $|y^\text{lb} - y^\text{ub}| \leq 5\sigma$, beyond which the gradient flattens and may lead to local minima.  
Alternatively, a convex penalty such as $\frac{1}{x^2}$ can be used, but introduces three challenges: (i) it is ill-posed near $x = 0$, requiring constraints like $|y_k - y_\mathcal{G}| > 0$; (ii) its steep gradient may destabilize optimization; (iii) it demands extensive tuning, with safety margins tied to penalty weights that are hard to generalize across scenarios.
\end{remark}

\subsection{Systematic Parameter Setting}
In \cref{eq:obj}, the two primary penalties are the deviation from reference and occupancy risk. While the ego vehicle is encouraged to follow the reference path, it must avoid collisions when sufficient space is available.  
Since the risk penalty is bounded by 1, we set $\lambda_\text{grid} = \bar{y}_\text{des}^2$ to match the deviation penalty at a maximum desired offset $\bar{y}_\text{des}$. For example, $\bar{y}_\text{des} = 10\,\text{m}$ ensures the vehicle does not deviate more than $10\,\text{m}$ from the reference path in the presence of a single occupancy\footnote{This guarantee applies to a given \textit{reference path} and a \textit{single} occupancy. For multiple occupancies, normalization is required.}.  
Additionally, the risk penalty should exceed the deviation penalty when the occupancy lies within the safety margin $\sigma$.
Formally, for $Q_d=I$ and $\forall\bar{y}\in[0,\sigma]$, 
\begin{align}
    \lambda_\text{grid} \exp{\left(-\frac{\bar{y}^2}{2(\sigma\cdot\tau)^2}\right)} \ge \bar{y}^2 \label{eq:risk_pen_ge_dev} \\
    \implies \lambda_\text{grid} \ge \sigma^2 \exp{ \left(\frac{1}{2\tau^2}\right)}.\label{eq:lam_grid}
\end{align}
\begin{remark}
This analysis extends to multiple occupancies, as the LHS in \cref{eq:risk_pen_ge_dev} grows monotonically with occupancy count, while the RHS remains constant
\end{remark}

We now show that the deviation penalty under the maximum steering angle does not exceed the risk penalty within the desired longitudinal distance $\sigma$. Using the circular arc model, the maximum lateral deviation is given by
$y_\text{max}(s) = R\left(1-\cos{\left(s/R\right)}\right)$,
where $s$ is the longitudinal distance, $L$ is the wheelbase, and $R = \frac{1}{\tan (\delta_\text{max})}$ is the turning radius.We want to prove that:
\begin{equation}
    \lambda_\text{grid} \exp{\left(-\frac{\bar{y}^2}{2(\sigma\cdot\tau)^2}\right)} \ge \bar{y}_\text{max}(s)^2 \;\;\forall s\in[0,\sigma].
\end{equation}
This holds if $\bar{y}\geq\bar{y}_{\text{max}}, \forall s\in[0,\sigma]$, assuming \cref{eq:lam_grid} holds.

\begin{proof}
Let $t = s/R \ge 0$ and $g(t) = t - (1-\cos t)$. Since $g'(t) = 1 - \sin t \ge 0$ and $g(0)=0$, it follows that $1-\cos t \le t, \forall t \ge 0$. Then, $\bar y_{\max}(s) = R\bigl(1-\cos(s/R)\bigr) = R(1-\cos t) \le Rt = s \implies \bar y_{\max}(s) \le s, \forall s\in[0,\sigma].$ 
\end{proof}

\section{Validation studies}\label{sec:validation}
\subsection{Simulation studies}
To validate ONRAP comprehensively, we conduct both qualitative and quantitative studies under reference noise, occupancy noise, and high-curvature scenarios. These experiments emulate real-world challenges, particularly perception and localization uncertainties. 

\begin{figure}
    \centering
    \includegraphics[width=1\columnwidth]{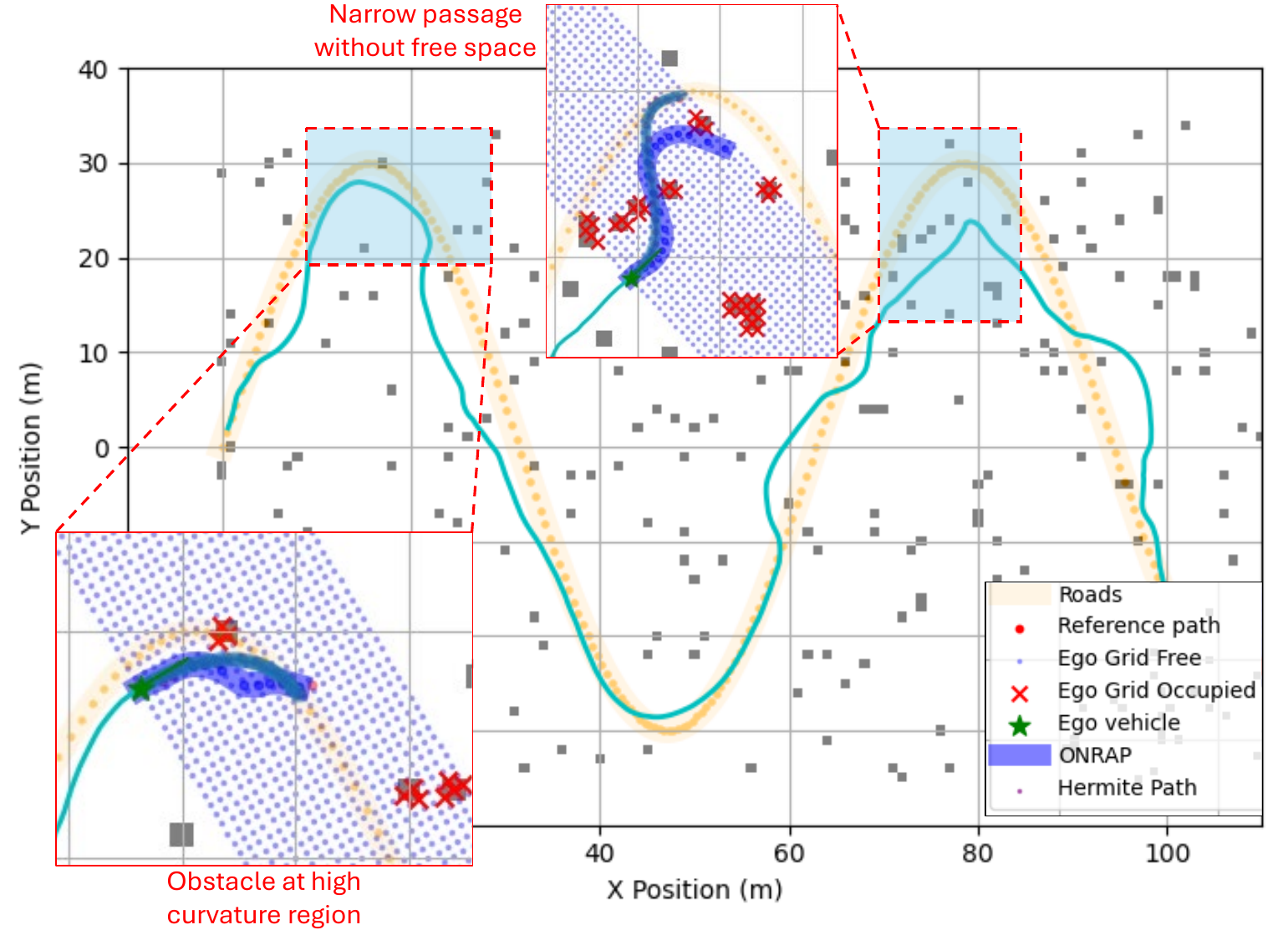}
    \caption{\textbf{Path planning results in the validation scenario.} The ego vehicle (green star) navigates a sinusoidal route (yellow) without access to ground-truth values. The magnified regions highlight two primary challenges, among others: (i) obstacles located at regions of high curvature, and (ii) a narrow passage between occupied cells with no direct free space along the route. The cyan line represents the simulated trajectory generated by the planner.}
    \label{fig:sim_result}
\end{figure}

In the scenario shown in \autoref{fig:sim_result}, the ego vehicle follows a sinusoidal route with high curvature and dense obstacles, using only a noisy reference path (red) and an occupancy map to emulate localization and perception uncertainty.  
A point from the noisy reference is selected as the local goal, and a Quintic Hermite path is generated as described in \autoref{sec:ref-generation}. While this path is smooth, bounded uniform spatial noise is injected to test robustness. Simulation parameters are listed in \autoref{tab:planner_params}.  
Both reference and occupancy noise are set to $0.3\,\text{m}$. Given the ego vehicle size of $2\,\text{m} \times 1\,\text{m}$, this corresponds to a $15\text{--}30\%$ localization displacement error, which is significant in cluttered environments.


\begin{table}[h]
  \centering
  \footnotesize
  \renewcommand{\arraystretch}{1.1}
  \setlength{\tabcolsep}{6pt}
  \begin{tabular*}{0.88\linewidth}{@{\extracolsep{\fill}} ll ll @{}}
    \toprule
    \textbf{Parameter} & \textbf{Value} & \textbf{Parameter} & \textbf{Value} \\
    \midrule
    Planning horizon & \(10~\mathrm{m}\)     & \(\lambda_{\text{curve}}\) & \(10\) \\
    \(d_s\)          & \(0.5~\mathrm{m}\)    & \(\lambda_{\text{grid}}\)  & \(10^{2}\) \\
    \(Q_d\)          & \(1\cdot I\)& \(\alpha_{\text{decay}}\)  & \(0.95\) \\
    \(Q_u\)          & \(1\cdot I\)          & \(\sigma\)                 & \(1.5\) \\
    \(u^{\min}\)     & \(-1~\mathrm{rad}\) & \(u^{\max}\)               & \(1~\mathrm{rad}\) \\
    \(\ell_{f},\ell_{r}\)     & \(1\)                 & \(\tau\) & \(2/3\) \\
    \bottomrule
  \end{tabular*}
  \vspace{5pt}
  \caption{Planner parameters.}
  \label{tab:planner_params}
\end{table}

\begin{figure}
    \centering
    \includegraphics[width=1\linewidth]{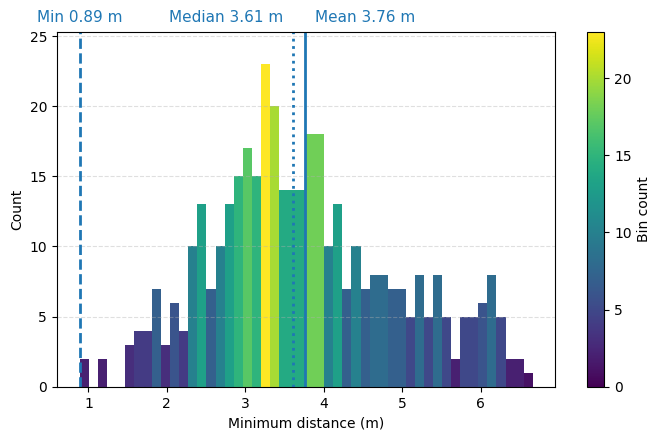}
    \caption{\textbf{Histogram of minimum distance to occupancy at each simulation step.} The global minimum occurs when the ego vehicle passes between two occupied cells, with a maximum spacing of $0.89\,\mathrm{m}$ during the event (thus increasing $\sigma$ does not improve clearance). ONRAP does not guarantee strict safety due to the absence of hard constraints; however, no collisions were observed empirically in this very cluttered scenario.}
    \label{fig:min_dist_hist}
\end{figure}

The planning performance of ONRAP is illustrated in \autoref{fig:sim_result}. We highlight three key observations:

\textbf{First}, ONRAP is able to find a path that balances between proximity to obstacles and adherence to the reference path when parameters are tuned to the ego vehicle size.
With the current parameter setting, the minimum lateral clearance is approximately $2\sigma$ (\autoref{fig:min_dist_hist}) when deviation from the reference is required. This suggests that $\sigma = \text{ego width} + \epsilon$, where $\epsilon$ is a safety buffer\footnote{Obstacle size need not be estimated, as it is encoded in the occupancy grid, so the desired safety margin is evaluated from an obstacle's edge.}, is a reasonable way to tune the parameters. This aligns with the analysis in \autoref{fig:risk_pareto}, where the risk with $\tau \approx 0.7$ degrades smoothly until $2\sigma$, helping achieve the desired distance around $2\sigma$.

\textbf{Second}, ONRAP is robust to bounded uniform noise, producing smooth trajectories under noisy perception and reference path. However, the deviation direction around an obstacle may oscillate due to compounded errors from perception noise, reference noise, and grid discretization. For example, the top block in \autoref{fig:sim_result} shows ONRAP initially planning to go between the two obstacles, but then adjusting to pass on the right at a later time.

\textbf{Third}, ONRAP does not formally guarantee collision avoidance. In cluttered spaces or under abrupt reference changes, minimum clearance can drop below half the vehicle width; thus, ONRAP should be paired with a downstream speed planner for longitudinal safety.

\begin{figure}
    \centering
    \includegraphics[width=1\linewidth]{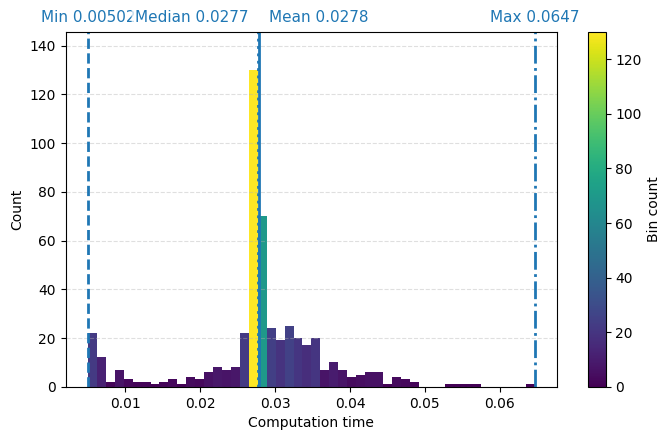}
    \caption{\textbf{Histogram of computation time.} Experiments are conducted on a 13th Gen Intel(R) Core(TM) i9-13900HX processor, which provides relatively high computational power. Efficiency results on an NVIDIA Jetson Orin Nano are reported in \autoref{sec:f1tenth}.}
    \label{fig:comp_time_hist}
\end{figure}

Overall, ONRAP reliably generates smooth paths without collisions or infeasibility. Computation time analysis in \autoref{fig:comp_time_hist}, with mean and maximum runtimes of $0.0278\,\text{s}$ and $0.0647\,\text{s}$ respectively, confirms real-time applicability.

\subsection{Quantitative Analysis}
We extend the validation study with Monte-Carlo simulations and compare the performance with widely-applied path planning benchmarks: A$^\star$ and RRT$^\star$. We modify the baselines’ terminal conditions at each time step: rather than requiring them to discover a collision-free local goal, we provide a pre-validated collision-free goal. 
This eliminates additional exploration, making their task easier. Although this favors the baselines, it lets us evaluate local planning efficiency cleanly and avoids frequent infeasible episodes that would otherwise interrupt runs.
Except for occupancy positions, all scenarios share identical settings: the same occupancy density, identical noise levels in both occupancy and reference path, and unchanged planning parameters as listed in \autoref{tab:planner_params}.

\autoref{tab:comparative-analysis} reports the results from the monte-carlo simulations with 100 randomly initialized scenarios. The evaluation metrics include: \{run-time\} to evaluate \textbf{efficiency}, \{success rate\footnote{A scenario is considered successful if the ego vehicle maintains a clearance of more than \texttt{ego\_width}/2 from all the obstacles.}, min/avg distance to the closest obstacle\} to evaluate \textbf{safety}, and \{max curvature, path length\} to evaluate \textbf{smoothness}. Note that the metrics are evaluated over the actual traversed trajectory, not the planned trajectory (as the ego movement follows the first point of each planned trajectory), and averaged over the 100 runs. The quantitative analysis was performed on a system running Ubuntu 22.04 LTS, equipped with an Intel® i7 CPU @ 2.40GHz and an NVIDIA RTX 3090 GPU.

From a broader perspective, ONRAP outperforms the baselines in efficiency, safety, and smoothness. It achieves a higher success rate despite baselines enforcing hard constraints on occupied regions and benefiting from collision-free goals, due to ONRAP’s superior noise handling. ONRAP also maintains greater obstacle clearance on average and in the worst case, corroborating its higher success rate.  
Furthermore, ONRAP delivers significantly smoother trajectories -- the maximum curvature is at least $4\times$ lower than that of the baselines, indicating improved comfort. While ONRAP’s paths are slightly longer due to deliberate safety-driven detours, baselines ignore vehicle kinematics and exploit unrealistic maneuvers (e.g., lateral jumps), resulting in shorter but infeasible paths.


\begin{table}[]
\centering
\caption{Comparative Analysis over 100 randomized scenarios}
\label{tab:comparative-analysis}
\resizebox{0.875\linewidth}{!}{%
\begin{tabular}{|
>{\columncolor[HTML]{FFD966}}c |c|c|c|}
\hline
\textbf{Method / Metric} &
  \cellcolor[HTML]{FD6864}\textbf{ONRAP} &
  \cellcolor[HTML]{D0D0D0}\textbf{A$^\star$} &
  \cellcolor[HTML]{D0D0D0}\textbf{RRT$^\star$} \\ \hline
\textbf{Run-time ($s$)} &
  \cellcolor[HTML]{DAF2D0}{\color[HTML]{242424} 0.033} &
  0.041 &
  0.215 \\ \hline
\textbf{Success Rate ($\%$)} &
  \cellcolor[HTML]{DAF2D0} 88 &
  37 &
  22 \\ \hline
\textbf{Min Dist ($m$)} &
  \cellcolor[HTML]{DAF2D0}{\color[HTML]{242424} 0.907} &
  {\color[HTML]{242424} 0.316} &
  {\color[HTML]{242424} 0.247} \\ \hline
\textbf{Avg Dist ($m$)} &
  \cellcolor[HTML]{DAF2D0}{\color[HTML]{242424} 3.994} &
  {\color[HTML]{242424} 3.062} &
  {\color[HTML]{242424} 2.695} \\ \hline
\textbf{Max Curv ($m^{-1}$)} &
  \cellcolor[HTML]{DAF2D0}{\color[HTML]{242424} 0.940} &
  {\color[HTML]{242424} 4.379} &
  {\color[HTML]{242424} 5.818} \\ \hline
\textbf{Path Len ($m$)} &
  {\color[HTML]{242424} 248.185} &
  \cellcolor[HTML]{DAF2D0}{\color[HTML]{242424} 205.556} &
  {\color[HTML]{242424} 228.875} \\ \hline
  
\end{tabular}
}
\end{table}

\subsection{F1TENTH experiments}    \label{sec:f1tenth}
We further demonstrate the performance of ONRAP on a 1/10th-scale F1TENTH racecar\footnote{Vehicle size: $0.5\,\text{m}$ length $\times 0.2\,\text{m}$ width} in an indoor office environment with narrow passages and diverse obstacles such as desks, chairs, bags, tables, and pedestrians. The test car is equipped with an NVIDIA Jetson Orin Nano (6-core Arm Cortex-A78AE, 8\,GB), which offers limited computational resources. Path planning runs concurrently with perception (Hokuyo 10LX LiDAR) and localization modules, requiring efficient real-time performance. On average, ONRAP runs at a computation rate of 10\,Hz alongside other modules.

\begin{figure}
    \centering
    \includegraphics[width=0.95\linewidth]{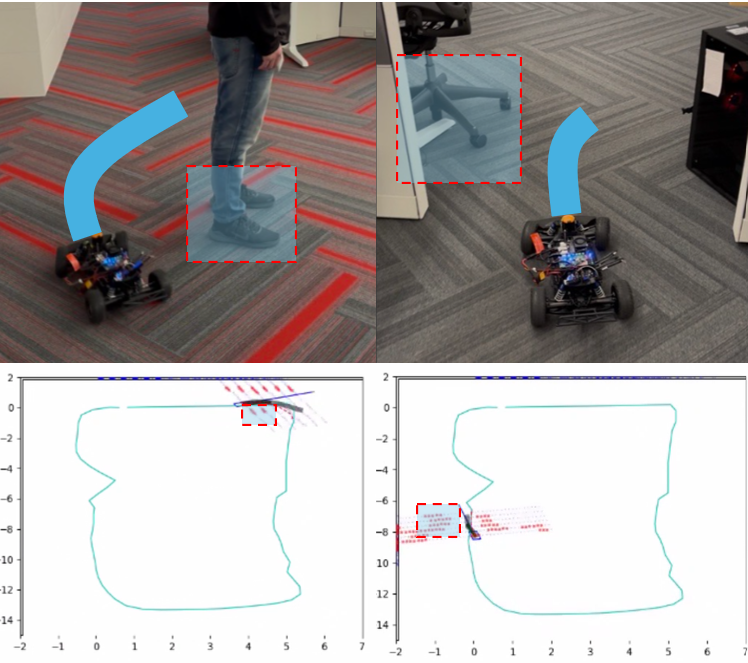}
    \caption{\textbf{F1TENTH experiment setup.} Experiments are conducted in an indoor office environment. The occupied regions are indicated by dashed red rectangles, and the ego vehicle’s planned path is shown in cyan.}
    \label{fig:rc_result}
    \vspace{-5pt}
\end{figure}

\begin{figure}
    \centering
    \includegraphics[width=0.95\linewidth]{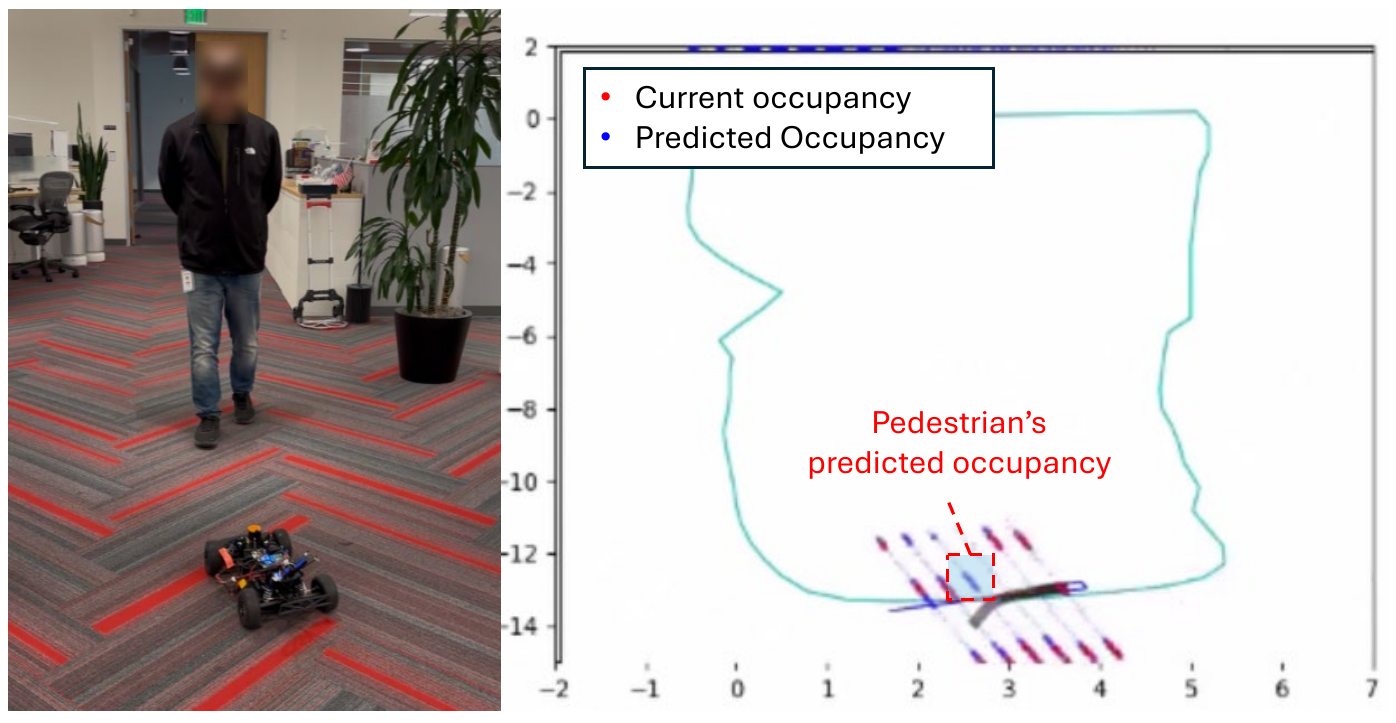}
    \caption{\textbf{F1TENTH experiment with a moving obstacle.} The future motion of an arbitrary dynamic obstacle is projected directly onto the occupancy grid, enabling the ego vehicle to anticipate and avoid the obstacle in advance.}
    \label{fig:rc_result_occ_pred}
    \vspace{-5pt}
\end{figure}

\autoref{fig:rc_result} shows two snapshots from the experiment: avoiding a stationary pedestrian and navigating a narrow passage between office equipment. During these trials, compounded localization and perception errors of approximately $\pm 0.5\,\text{m}$, about $250\%$ of the vehicle width, were observed. Despite this, ONRAP consistently completed the loop without collisions in static scenarios, even under high noise levels.  
Failures occurred with dynamic obstacles, particularly when they approached rapidly or forced the ego vehicle toward other occupancies. In such cases, a speed planner is necessary to maintain safe distances; we used a fixed speed of $0.5\,\text{m/s}$, but recovery strategies (e.g., \cite{tariq2023rcms,baldini2024don}) must be employed. When pedestrians moved predictably (e.g., walking straight), ONRAP successfully generated appropriate deviation plans (\autoref{fig:rc_result_occ_pred}).

\begin{remark}
We used \textbf{identical penalty weights} in both simulation and F1TENTH experiments: $Q_d$, $Q_u$, $u^\text{min}$, $u^\text{max}$, $\lambda_\text{curve}$, $\lambda_\text{grid}$, $\alpha_\text{decay}$, and $\tau$ remained unchanged from \autoref{tab:planner_params}. Vehicle-specific parameters, such as size and safety distance $\sigma$, were updated for F1TENTH (set to $0.5\,\text{m}$). This demonstrates ONRAP’s strong potential for generalizability across platforms and environments.
\end{remark}


\subsection{Discussion}\label{sec:discussion}
\textbf{Usefulness and Applications:} ONRAP remains effective under compounded perception and localization errors, as well as noisy local goals, without relying on semantic segmentation or object recognition. In practice, it delivers kinematically feasible, smooth trajectories in real time and transfers across platforms with minimal re-tuning (weights as in Table~I; only vehicle footprint and $\sigma$ adapted for F1TENTH).  
ONRAP offers broad applicability in robotics, serving as:  
(i) a primary path planner in mobility stacks operating in resource-constrained high-noise environments;  
(ii) an easy-to-implement research baseline that balances interpretability and practicality; and  
(iii) a model to support learning-based methods during training, e.g., model-based guidance to accelerate learning \cite{huang2020model,morgan2021model,wang2023combined}.

\textbf{Limitations:} ONRAP does not guarantee safety, as it relies on soft constraints for collision avoidance and behaves similarly to other non-hard-constrained approaches. Being based on a path–speed decomposition scheme, it requires a downstream speed planner \cite{anon2024multi,isele2025delayed} to maintain longitudinal safety.  
Additionally, ONRAP operates without object recognition, meaning obstacle dynamics depend solely on occupancy-flow prediction; incorporating partial semantic information could improve prediction accuracy and restrict drivable areas, offering a better balance between practicality and precision.

\section{Conclusion} \label{sec:conclusion}
We present ONRAP, a noise-resilient, occupancy-driven path-planning module that formulates local motion generation as a spatial-domain nonlinear program. By grounding collision avoidance in occupancy and flow fields, ONRAP achieves robustness to perception and localization noise while maintaining kinematic feasibility and real-time performance. Simulations and F1TENTH experiments confirm smooth, safe behavior in cluttered, dynamic environments. Future work will couple ONRAP with speed planning and learned occupancy-flow predictors to strengthen safety and interactive-agent handling.

\bibliographystyle{IEEEtran}
\bibliography{refs} 

@inproceedings{tariq2022slas,
  title={Slas: Speed and lane advisory system for highway navigation},
  author={Tariq, Faizan M and Isele, David and Baras, John S and Bae, Sangjae},
  booktitle={2022 IEEE 61st Conference on Decision and Control (CDC)},
  pages={6979--6986},
  year={2022},
  organization={IEEE}
}

@inproceedings{tariq2023rcms,
  title={Rcms: Risk-aware crash mitigation system for autonomous vehicles},
  author={Tariq, Faizan M and Isele, David and Baras, John S and Bae, Sangjae},
  booktitle={2023 IEEE 26th International Conference on Intelligent Transportation Systems (ITSC)},
  pages={3950--3957},
  year={2023},
  organization={IEEE}
}

@article{dougherty1989nonnegativity,
  title={Nonnegativity-, monotonicity-, or convexity-preserving cubic and quintic Hermite interpolation},
  author={Dougherty, Randall L and Edelman, Alan S and Hyman, James M},
  journal={mathematics of computation},
  volume={52},
  number={186},
  pages={471--494},
  year={1989}}

@article{ionut2022spherical,
  title={On the spherical clothoid},
  author={Ionut, Alexandru},
  journal={arXiv preprint arXiv:2203.07963},
  year={2022}
}

@inproceedings{ryu2025iann,
  author = {Ryu, Kanghyun and Sung, Minjun and Gupta, Piyush and D'sa, Jovin and Tariq, Faizan M. and Isele, David and Bae, Sangjae},
  title = {IANN-MPPI: Interaction-Aware Neural Network-Enhanced Model Predictive Path Integral Approach for Autonomous Driving},
  booktitle = {2025 IEEE 28th International Conference on Intelligent Transportation Systems (ITSC)},
  year = {2025},
  organization={IEEE}
}

@INPROCEEDINGS{tariq2025frenet,
  author={Tariq, Faizan M. and Yeh, Zheng-Hang and Singh, Avinash and Isele, David and Bae, Sangjae},
  booktitle={2025 IEEE Intelligent Vehicles Symposium (IV)}, 
  title={Frenet Corridor Planner: An Optimal Local Path Planning Framework for Autonomous Driving}, 
  year={2025},
  volume={},
  number={},
  pages={686-693},
  doi={10.1109/IV64158.2025.11097649}}

@inproceedings{baldini2024don,
  title={Don't Get Stuck: A Deadlock Recovery Approach},
  author={Baldini, Francesca and Tariq, Faizan M and Bae, Sangjae and Isele, David},
  booktitle={2024 IEEE 27th International Conference on Intelligent Transportation Systems (ITSC)},
  pages={3688--3695},
  year={2024},
  organization={IEEE}
}

@inproceedings{huang2020model,
  title={Model-based or model-free, a review of approaches in reinforcement learning},
  author={Huang, Qingyan},
  booktitle={2020 International Conference on Computing and Data Science (CDS)},
  pages={219--221},
  year={2020},
  organization={IEEE}
}

@inproceedings{morgan2021model,
  title={Model predictive actor-critic: Accelerating robot skill acquisition with deep reinforcement learning},
  author={Morgan, Andrew S and Nandha, Daljeet and Chalvatzaki, Georgia and D’Eramo, Carlo and Dollar, Aaron M and Peters, Jan},
  booktitle={2021 IEEE International Conference on Robotics and Automation (ICRA)},
  pages={6672--6678},
  year={2021},
  organization={IEEE}
}

@article{wang2023combined,
  title={A combined reinforcement learning and model predictive control for car-following maneuver of autonomous vehicles},
  author={Wang, Liwen and Yang, Shuo and Yuan, Kang and Huang, Yanjun and Chen, Hong},
  journal={Chinese Journal of Mechanical Engineering},
  volume={36},
  number={1},
  pages={80},
  year={2023},
  publisher={Springer}
}

@inproceedings{anon2024multi,
  title={Multi-profile quadratic programming (mpqp) for optimal gap selection and speed planning of autonomous driving},
  author={Anon, Alexandre Miranda and Bae, Sangjae and Saroya, Manish and Isele, David},
  booktitle={2024 IEEE International Conference on Robotics and Automation (ICRA)},
  pages={12158--12164},
  year={2024},
  organization={IEEE}
}

@article{mahjourian2022occupancy,
  title={Occupancy flow fields for motion forecasting in autonomous driving},
  author={Mahjourian, Reza and Kim, Jinkyu and Chai, Yuning and Tan, Mingxing and Sapp, Ben and Anguelov, Dragomir},
  journal={IEEE Robotics and Automation Letters},
  volume={7},
  number={2},
  pages={5639--5646},
  year={2022},
  publisher={IEEE}
}

@inproceedings{agro2023implicit,
  title={Implicit occupancy flow fields for perception and prediction in self-driving},
  author={Agro, Ben and Sykora, Quinlan and Casas, Sergio and Urtasun, Raquel},
  booktitle={Proceedings of the IEEE/CVF conference on computer vision and pattern recognition},
  pages={1379--1388},
  year={2023}
}

@inproceedings{hu2021fiery,
  title={Fiery: Future instance prediction in bird's-eye view from surround monocular cameras},
  author={Hu, Anthony and Murez, Zak and Mohan, Nikhil and Dudas, Sof{\'\i}a and Hawke, Jeffrey and Badrinarayanan, Vijay and Cipolla, Roberto and Kendall, Alex},
  booktitle={Proceedings of the IEEE/CVF International Conference on Computer Vision},
  pages={15273--15282},
  year={2021}
}

@article{bansal2019chauffeurnet,
  title={ChauffeurNet: Learning to Drive by Imitating the Best and Synthesizing the Worst. Robotics: Science \& Systems (RSS), art},
  author={Bansal, Mayank and Krizhevsky, Alex and Ogale, Abhijit},
  journal={arXiv preprint arXiv:1812.03079},
  year={2019}
}

@inproceedings{koenig2002dlite,
  title={D* lite},
  author={Koenig, Sven and Likhachev, Maxim},
  booktitle={Eighteenth national conference on Artificial intelligence},
  pages={476--483},
  year={2002}
}

@article{dolgov2008hybrida,
  title={Practical search techniques in path planning for autonomous driving},
  author={Dolgov, Dmitri and Thrun, Sebastian and Montemerlo, Michael and Diebel, James},
  journal={ann arbor},
  volume={1001},
  number={48105},
  pages={18--80},
  year={2008}
}

@INPROCEEDINGS{isele2025delayed,
  author={Isele, David and Añon, Alexandre Miranda and Tariq, Faizan M. and Yeh, Goro and Singh, Avinash and Bae, Sangjae},
  booktitle={2025 IEEE International Conference on Robotics and Automation (ICRA)}, 
  title={Delayed-Decision Motion Planning in the Presence of Multiple Predictions}, 
  year={2025},
  volume={},
  number={},
  pages={11743-11749},
  doi={10.1109/ICRA55743.2025.11128178}}

@article{fleuret2008multicamera,
  title={Multicamera people tracking with a probabilistic occupancy map},
  author={Fleuret, Francois and Berclaz, Jerome and Lengagne, Richard and Fua, Pascal},
  journal={IEEE transactions on pattern analysis and machine intelligence},
  volume={30},
  number={2},
  pages={267--282},
  year={2008},
  publisher={IEEE}
}

@inproceedings{stentz1995focussed,
  title={The focussed d\^{}* algorithm for real-time replanning},
  author={Stentz, Anthony and others},
  booktitle={IJCAI},
  volume={95},
  pages={1652--1659},
  year={1995}
}

@article{sadigh2018planning,
  title={Planning for cars that coordinate with people: leveraging effects on human actions for planning and active information gathering over human internal state},
  author={Sadigh, Dorsa and Landolfi, Nick and Sastry, Shankar S and Seshia, Sanjit A and Dragan, Anca D},
  journal={Autonomous Robots},
  volume={42},
  number={7},
  pages={1405--1426},
  year={2018},
  publisher={Springer}
}

@inproceedings{kong2015kinematic,
  title={Kinematic and dynamic vehicle models for autonomous driving control design},
  author={Kong, Jason and Pfeiffer, Mark and Schildbach, Georg and Borrelli, Francesco},
  booktitle={2015 IEEE intelligent vehicles symposium (IV)},
  pages={1094--1099},
  year={2015},
  organization={IEEE}
}

@article{o2020f1tenth,
  title={F1tenth: An open-source evaluation environment for continuous control and reinforcement learning},
  author={O'Kelly, Matthew and Zheng, Hongrui and Karthik, Dhruv and Mangharam, Rahul},
  journal={Proceedings of Machine Learning Research},
  volume={123},
  year={2020}
}

@inproceedings{niemeyer2019occupancy,
  title={Occupancy flow: 4d reconstruction by learning particle dynamics},
  author={Niemeyer, Michael and Mescheder, Lars and Oechsle, Michael and Geiger, Andreas},
  booktitle={Proceedings of the IEEE/CVF international conference on computer vision},
  pages={5379--5389},
  year={2019}
}
\end{document}